\documentclass[letterpaper, 10 pt,conference]{ieeeconf}

\IEEEoverridecommandlockouts
\usepackage{flushend}
\usepackage{amsmath}
\usepackage{array}
\usepackage{cite} 
\usepackage{epsfig}
\usepackage{chngcntr}
\usepackage{amssymb}
\usepackage{amsmath}
\usepackage{graphicx}
\usepackage{verbatim}
\usepackage[caption = false]{subfig}
\usepackage{float}
\usepackage{booktabs}
\usepackage{multirow}
\usepackage{tabularx}
\usepackage{threeparttable}
\usepackage{hyperref}

\usepackage[lined,ruled,commentsnumbered, linesnumbered]{algorithm2e}
\usepackage{algpseudocode}
   \graphicspath{{Figures/}}
   \DeclareGraphicsExtensions{.pdf,.jpg,.png,.eps}

\usepackage[text={7in,9.5in},centering]{geometry}

\title{\LARGE \bf Situation-aware Autonomous Driving  Decision Making  \\ with Cooperative Perception on Demand }

\author{ Wei Liu
\thanks{Wei Liu is with Department of Mechanical Engineering, National University of Singapore, Kent Ridge, Singapore \tt\small{liu\_wei}@ u.nus.edu}
\thanks{This research was supported by the Future Urban Mobility project of the Singapore-MIT Alliance for Research and Technology (SMART) Center, with funding from Singapore's National Research Foundation.}
}
\begin{document}
\maketitle
\thispagestyle{empty}
\pagestyle{empty}

\begin{abstract}
 This paper investigates the impact of cooperative perception on autonomous driving decision making on urban roads. The extended perception range contributed by the cooperative perception can be properly leveraged to address the implicit dependencies within the vehicles, thereby the vehicle decision making performance can be improved. Meanwhile, we acknowledge the inherent limitation of wireless communication and propose a Cooperative Perception on Demand (CPoD) strategy, where the cooperative perception will only be activated when the extended perception range is necessary for proper situation-awareness. The situation-aware decision making with CPoD is modeled as a Partially Observable Markov Decision Process (POMDP) and solved in an online manner. The evaluation results demonstrate that the proposed approach can function safely and efficiently for autonomous driving on urban roads.
\end{abstract}

\section{Introduction}\label{sec:Introduction}
As autonomous vehicles being navigating on the urban road, the proper decision making requires the comprehensive awareness of the surrounding environment, which includes both the road context and the other vehicles' motion intention. The knowledge of the road context can help to narrow down the reasoning scope of the other traffic participants' behaviors \cite{sivaraman2013observing}. Similarly, to understand the semantic meaning of the other vehicles' behaviors, i.e., motion intention, can contribute a more robust prediction of their future motion, which in turn can be employed for a more accurate risk evaluation\cite{bandyopadhyay2013intention}.

The difficulty of situation awareness, however, is always aggravated by the limited autonomous vehicle perception ability, which has to experience the difficulties introduced by the inevitable sensor noise and sensing occlusion. The resulting sensing limitation can dramatically weaken the autonomous vehicle's ability of perceiving the obstacles' states and blind the implicit behavior dependencies within the vehicles. As a preeminent candidate to extend perception range without substantial additional costs, cooperative perception is the exchange of local sensing information with other vehicles or infrastructures via wireless communications \cite{li2012cooperative}. The perception range thereby can be considerably extended up to the boundary to the connected vehicle, which thereby can be useful for better situation awareness. 

While the benefits of cooperative perception has been well recognized, there always come with some challenges of the cooperative perception applications on autonomous driving \cite{kim2014multivehicle}. Firstly, the challenge can arise from the balance of information quantity and quality. There still lacks the comprehensive strategy and general rules to efficiently and robustly process and represent the information shared by the different participants. Secondly, the perception uncertainties might increase dramatically according to the growth of the extended sensing range due to the perception processing error, which hints that the perception failures need to be explicitly considered. Last but not least, the applicability of cooperative perception can be always obstructed by the inherent communication constraints, such as the limited bandwidth. 

In this paper, we investigate the impact of cooperative perception on situation-aware decision making for autonomous driving on urban roads. Extending upon our previous work in \cite{liu2015situation}, we explicitly consider a sensing occlusion problem and aim at employing cooperative perception to address this issue. Given the fact that the communication between vehicles is always constrained, we propose a Cooperative Perception on Demand (CPoD) strategy, which means that the cooperative perception will be activated only when the on-board sensing is not informative enough for proper situation awareness. To properly feature CPoD inside the situation-aware decision making module, the Partially Observable Markov Decision Making Process (POMDP) is utilized for its innate ability to balance the exploration and exploitation. The evaluation results show that the proposed algorithm is efficient and safe enough to handle some common urban driving scenarios.


Compared to the previous work, the contributions of this study are obvious as follows: 
\begin{itemize}
\item We investigate the impact of cooperative perception on autonomous vehicle decision making module. 
\item The proposal of CPoD can actively alleviate the constraints on vehicle communication. 
\end{itemize}

The remainder of this paper is organized as follows. Section \ref{sec:Related_Work} introduces some related works and Section \ref{sec:Preliminaries} provides the preliminaries on both cooperative perception and POMDP. The problem statement is discussed in Section \ref{sec:Problem_Statement}. Thereafter, the detailed discussion of the situation-aware decision making with CPoD is presented in Section \ref{sec:Decision_Making}. The evaluations of the proposed algorithm are elaborated in Section \ref{sec:Evaluation} and this paper is concluded in Section \ref{sec:Conclusion}.


\section{Related Work}\label{sec:Related_Work}
Three folds of literature directly related to our study will be identified. Firstly, the application of cooperative perception on autonomous driving will be briefly discussed. The second one is the multi-agent active sensing using decision-theoretic techniques, and the last one is how the POMDP has been employed for autonomous driving decision making.

Given the obvious benefits of cooperative perception, the extension of cooperative perception to autonomous driving has attracted dramatic research interests recently. Given the extended perception range, the motion planning horizon can be extended to the boundary of the connected vehicles, which thereby can contribute to an earlier hidden obstacle avoidance \cite{liu2013motion}. Similarly, the situation awareness can be improved as well, because each vehicle's behavior is correlated to others, which hints that the expanded sensing information through cooperative perception can be useful for better reasoning of the vehicles' behaviors \cite{liu2014vehicle}. 


The multi-agent active sensing using decision-theoretic techniques aims at properly coordinating the sensors or mobile platforms to achieve a wider sensing coverage or to reduce the uncertainties of object characteristics identification \cite{spaan2009decision,spaan2008cooperative}. While great achievements have been made, the communication constraint, however, is occasionally overlooked \cite{otte2014any}. Moreover, while these approaches can properly handle the planning for perception issue, but not vice versa. In a sense, the investigation on how the enhanced perception can contribute to the planning is not comprehensive.

Regarding the autonomous driving decision making, the most common approach is to manually tailor the specific action sets for different situations using a Finite State Machine or similar frameworks like behavior trees \cite{ferguson2008motion}. These approaches, however, lack the comprehensive understanding of the environment. The absence of full situational awareness can make the driving decisions become danger-prone, which has been evidenced by an incident of DUC \cite{fletcher2008cornell}. As an principle approach of balancing exploration and exploitation, POMDP has been widely adopted for autonomous driving decision making recently \cite{ulbrich2013probabilistic,bai2015intention,shaojun2014online}. The POMDP applications around the literature can vary from designing some specific driving behaviors, such as lane changing and merging\cite{ulbrich2013probabilistic}, to a more general traffic negotiation \cite{liu2015situation}.


As reviewed above, a general situation-aware decision making approach to integrate the above three components is still an interesting and challenging problem.


\section{Preliminaries}\label{sec:Preliminaries}

\subsection{Cooperative Perception}

In contrast to cooperative driving, where the vehicles are centrally controlled or share their motion intention directly, cooperative perception targets at sharing local sensing information only, which is more applicable to different traffic participants in typical urban traffic situations.

In the context of autonomous vehicle perception, the sensing information is typically projected into a local frame, which can be processed as a local observation that consists of a set of perceived obstacles or features representing the environment. Let $Z^{[i]}$ depict a local sensing map of the vehicle $i$, and let $\varphi^{[i]} = \{i-N_f,\cdots, i-1, i+1, \cdots, i+N_l \}$ be a string that defines the neighbors of a vehicle $i$, where $N_f, N_l$ denotes the size of the following and leading neighbors respectively. Then cooperative perception is formulated as,
\begin{equation}
\mathbf{Z}^{[i]} = Z^{[i]} \bigcup_{j\in\varphi^{[i]}} Z^{[j]}\otimes Q^{[j,i]},
\label{Map_Merge_Init}
\end{equation}
where $Q^{[j,i]}=[\tau^{[j,i]},\theta^{[j,i]}]$ denotes the relative pose of vehicle $j$ w.r.t. vehicle $i$, which consists of the translation $\tau^{[j,i]}$ and rotation $\theta^{[j,i]}$. Given the relative pose, the operator $\otimes$ computes the pose transformation. The operation $\bigcup$ thereby is called \textit{map merging} and the resulting $\mathbf{Z}^{[i]}$ denotes the extended sensing map of vehicle $i$ using cooperative perception. Since the agents participated in the cooperative perception are designed to work in a decentralized manner, we adhere to the following rules in this study:
\begin{itemize}
	\item There is no central entity required for the operation.
	\item There is no common communication facility, which means only local point-to-point communication between neighbors is considered.
	\item There is no communication delay being considered.
\end{itemize} 

\subsection{Partially Observable Markov Decision Process}

A POMDP is formally a tuple $\{\mathcal{S},\mathcal{A},\mathcal{Z},T,O,R,\gamma\}$, where $\mathcal{S}$ is the state space, $\mathcal{A}$ is a set of actions and $\mathcal{Z}$ denotes the observation space. The transition function $T(s',s,a)=\mathrm{Pr}(s'|s,a):\mathcal{S}\times\mathcal{A}\times\mathcal{S}$ models the probability of transiting to state $s'\in\mathcal{S}$ when the agent takes an action $a\in \mathcal{A}$ at state $s\in\mathcal{S}$. The observation function $O(z,s',a)=\mathrm{Pr}(z|s',a):\mathcal{S}\times\mathcal{A}\times\mathcal{Z}$, similarly, gives the probability of observing $z\in\mathcal{Z}$ when action $a\in\mathcal{A}$ is applied and the resulting state is $s'\in\mathcal{S}$. The reward function $R(s,a):\mathcal{S}\times\mathcal{A}$ is the reward obtained by performing action $a\in\mathcal{A}$ in state $s\in\mathcal{S}$, and $\gamma \in [0, 1)$ is a discount factor.


The solution to the POMDP thereby is an optimal policy $\pi^{*}$ that maximizes the expected accumulated reward $\mathbb{E}(\sum_{t=0}^{\infty}\gamma^{t}R(a_{t},s_{t}))$, where $s_{t}$ and $a_{t}$ denote the agent's state and action at time $t$, respectively. The true state, however, is not fully observable to the agent, thus the agent maintains a belief state $b\in\mathcal{B}$, i.e. a probability distribution over $\mathcal{S}$, instead. The policy $\pi:\mathcal{B}\rightarrow\mathcal{A}$ therefore maps a prescribed action $a\in\mathcal{A}$ from a belief $b\in\mathcal{B}$. 

\section{Problem Statement}\label{sec:Problem_Statement}

\begin{figure}
	\begin{center}
		\includegraphics[width=3.4in]{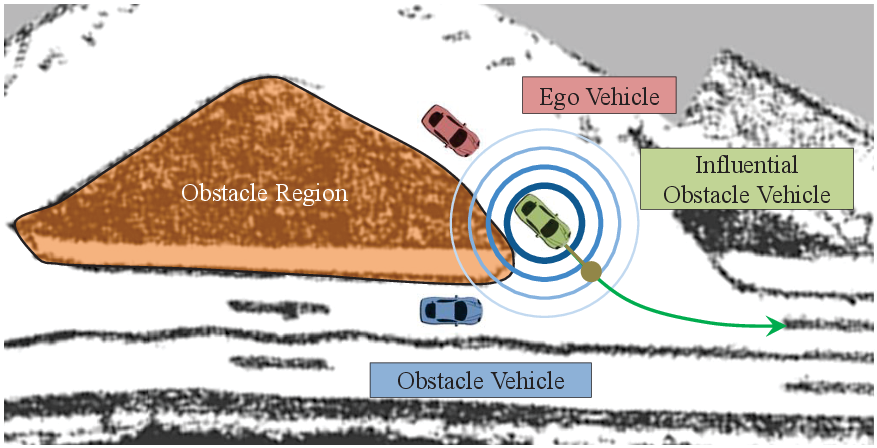}
		\caption{Illustrative example of CPoD: The blue obstacle vehicle is blind the red ego vehicle. Given the shared observation from the IOV, ego vehicle can conduct a more comprehensive situation reasoning. }
		\label{fig:CPoD_Explaination}
	\end{center}
\end{figure}

Onwards, we name the autonomous vehicle as an ego vehicle and consider all the other vehicles as obstacle vehicles. Also, we define the \textit{Influential Obstacle Vehicle} (IOV) as the obstacle vehicle that imposes the most significant impact on the ego vehicle's driving decisions, such as the leading vehicle on the single-lane road, or the vehicles to be negotiated at the intersection. 

In sight of the influence of IOV, the perception sharing between ego vehicle and IOV becomes more essential, where a more conscious decision can be made \emph{if an ego vehicle can see what IOV see}. Taking Fig. \ref{fig:CPoD_Explaination} for instance, the ego vehicle is following the leading vehicle to conduct lane merging. Since the ego vehicle's driving decision is heavily affected by the leading vehicle (identified as IOV), the inference over the leading vehicle's motion intention becomes essential. Without cooperative perception, another coming obstacle vehicle (blue) is blind to the ego vehicle. Given the on-board observation that the IOV is driven in a normal speed, the ego vehicle might draw a confident but danger-prone conclusion that the IOV will merge into the lane directly. On the other hand, if the IOV shares what it sees with the ego vehicle, the ego vehicle can then properly acknowledge the \emph{dependencies} between the IOV and the coming obstacle vehicle, such that a more reasonable inference can be made as: the IOV may have a fairly high chance to commit a deceleration.


Obviously, the perception sharing can contribute to a greatly extended perception range, and more importantly a comprehensive understanding of the situation evolution, but it does bring itself with some limitations, especially the communication constraints. Therefore, to properly identify the situation in which perception sharing should be activated can be quite helpful in sense. The difficulty of situation identification, however, can always be aggravated by the fact that the situation, including the obstacle vehicle behavior and environment context, is not fully observable. In sight of this, this paper proposes a \textit{Cooperative Perception on Demand} (CPoD) strategy and seeks to feature the CPoD inside the decision making module. In a sense, perception sharing will only be activated when the situation awareness becomes difficult, such as the IOV's motion intention is ambiguous or the collision risk is critical. 

Formally speaking, let $Z^{[e]}$ and $Z^{[iov]}$ denotes the sensing map of ego vehicle and IOV respectively, both of which consist of a set of environment features. As a function of the CPoD decision $a_{\text{\tiny CPoD}}\in\{\textsc{Active},\textsc{Deactive}\}$, the ego vehicle's extended sensing map $\mathbf{Z}^{[e]}$ is modeled as,
\begin{equation}
\mathbf{Z}^{[e]}(a_{\text{\tiny CPoD}}) = 
\begin{cases}
Z^{[e]} \bigcup Z^{[iov]}\otimes Q^{[iov,e]}, \textrm{ if } a_{\text{\tiny CPoD}} = \textsc{Active}, \\
Z^{[e]}, \textrm{ if } a_{\text{\tiny CPoD}} = \textsc{Deactive}.
\end{cases}
\end{equation}
As such, once the CPoD is activated, the ego vehicle's perception range can be extended to as far as IOV can reach. Thereafter, this study seeks to solve a decision making problem to minimize the cost expectation as,
\begin{equation}
\pi^{*} = \arg \min_{a_{t}\in\mathcal{A}}\mathbb{E}_{s_{t}\in\mathcal{S}}\{\sum_{t=0}^{\infty}\mathcal{L}(a_{t},s_{t})\mathrm{Pr}(s_{t}|\mathbf{Z}^{[e]}_{t}(a_{t}))\},
\end{equation}
where $s$ and $\mathcal{S}$ depict the situation state and state space respectively. The situation state $s$ is usually associated with certain uncertainties, which can include, to name a few, the road context, the obstacle vehicle's moving intention, etc. The inference over the situation state, thereby, is driven by the sensing information $\mathbf{Z}^{[e]}$. Moreover, $a =[a_{\text{\tiny CPoD}}, a_{\text{\tiny ACC}}]$ represents the ego vehicle's action, which consists of both the CPoD decision $a_{\text{\tiny CPoD}}$ and the acceleration command $a_{\text{\tiny ACC}}$, and $\mathcal{A}$ denotes the corresponding action space. The cost function $\mathcal{L}(a,s):\mathcal{A}\times\mathcal{S}\rightarrow\mathbb{R}^{+}$ targets to encourage the driving efficiency and penalize both the colliding risk and the vehicle communication, which thereby can properly balance the driving safety and information gaining. 

Worth mentioning that the sensing fusion and perception uncertainty modeling problems are out of the scope of this paper, the reader can refer to \cite{kim2014multivehicle} and \cite{xiaotong2014dars} for a more detailed discussion. In this context, we assume that a well-processed perception, especially the uncertainty modeling, is alway available. 

\section{Situation-aware Decision Making with CPoD}\label{sec:Decision_Making}

\subsection{Overview}\label{sec:System_Overview}

\begin{figure}
	\begin{center}
		\includegraphics[width=3.4in]{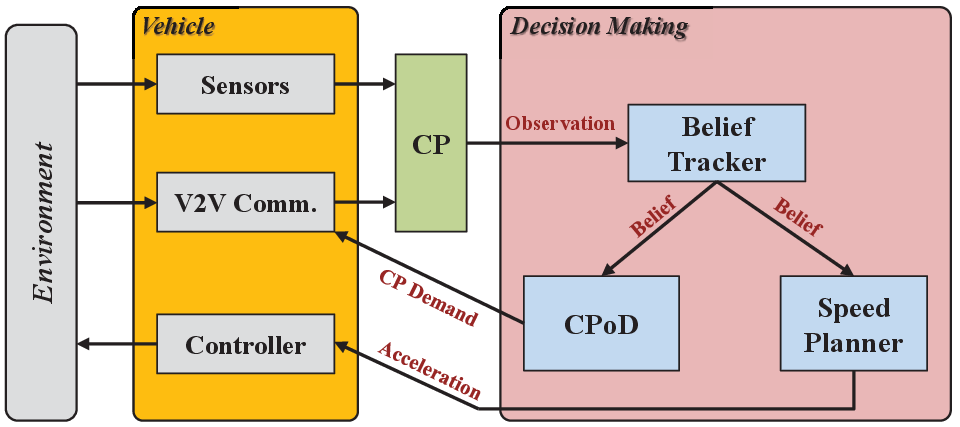}
		\caption{System framework for situation-aware decision making with CPoD. }
		\label{fig:System_Overview}
	\end{center}
\end{figure}

The overall framework of the autonomous driving decision making with CPoD is depicted in Fig. \ref{fig:System_Overview}. The autonomous vehicle is featured with the on-board sensors and the V2V communication ability, where the V2V communication will only be enabled when CPoD has been activated. Once the V2V communication is enabled, the cooperative perception module will start to process the sensing data provided by both the on-board sensors and the IOV, otherwise only the on-board sensing data will be processed. 

Given the augmented sensing information provided by cooperative perception, the decision making module will come into act. The decision making module is in change of making two sub-decisions: 1) decision about the CPoD and 2) decision about the ego vehicle's acceleration. Both of these decisions are made upon the situation state belief provided by the belief tracker.

\subsection{POMDP Modeling}
Hereafter, the POMDP model for situation-aware decision making with CPoD will be discussed. Generally speaking, the ego vehicle will receive the IOV's sensing information once the CPoD is activated, thereafter the acceleration command will be chosen, which drives the ego vehicle's state transition. Regarding the obstacle vehicles, their state transitions are driven by the inferred motion intention, the road context, and the implicit dependencies within the vehicles.

\subsubsection{Ego Vehicle State Transition Model}
The ego vehicle state $s^{[e]}$ consists of the vehicle pose $[\rm x^{[e]},\rm y^{[e]}, \theta^{[e]}]^{\rm T}$ and velocity $\rm v^{[e]}$. Given our previous discussion, the ego vehicle action space is discretized as,
\begin{gather}
\nonumber
\mathcal{A}=\mathcal{A}_{\text{CPoD}}\times\mathcal{A}_{\text{\tiny ACC}},
\\
\nonumber
\mathcal{A}_{\text{CPoD}} = \{\textsc{Active,Deactive}\},  \\\mathcal{A}_{\text{\tiny ACC}} = \{\textsc{Accelerate, Decelerate, Constant}\},
\end{gather}
which has been verified to be qualified for an ego vehicle to achieve most tactical maneuvers, such as traffic negotiation at intersections, maintaining the safe distance with other obstacle vehicles, etc.\cite{liu2015situation}. The vehicle steering control, on the other hand, is accomplished by following a pre-specified reference route using a close-loop path tracking algorithm. Given the acceleration command and steering angle, the ego vehicle system can be forwarded for a fixed time duration $\Delta t$ using the kinematic car model in \cite{lavalle2006planning}.


\subsubsection{Obstacle Vehicle State Transition Model}
Due to the on-board sensing limitation, some obstacle vehicles cannot be visible to the ego vehicle, thus we define $\varphi_{\text{vis}}$ as the set of obstacle vehicles that are within the range of the extended sensing map $\mathbf{Z}^{[e]}$. Each obstacle vehicle is associated with an unique identification number, and the state of obstacle vehicle $i$ is modeled as $s^{[i]} = [\rm x^{[i]}, \rm y^{[i]}, \theta^{[i]}, \rm v^{[i]}, \iota^{[i]}]^{\rm T}, i\in\varphi_{\text{vis}}$, where $\iota \in \mathcal{I}$ denotes the motion intention and $\mathcal{I}$ represents the motion intention space. Compared to the ego vehicle state, the inclusion of motion intention here is to provide a semantic meaning of the obstacle vehicle behavior and to properly model the obstacle vehicle's state transition. 

Rather than addressing the motion intention as hidden goals \cite{shaojun2014online,bandyopadhyay2013intention,bai2015intention}, the \textit{reaction} based motion intention is employed in this study \cite{liu2015situation}, which is abstracted from human driving behaviors by generalizing a bunch of vehicle behavior training data \cite{liu2015roadrule}. More specifically, given the vehicle behavior training data, the road context $\mathcal{C}$ can be learned first to represent the traffic rules and the typical vehicle motion pattern, thereafter the reaction is modeled as the deviation of the observed vehicle behavior from the generalized road context. Given the observed reaction, the motion intention space $\mathcal{I}$ is defined as $\mathcal{I} = \{Stopping,Hesitating,Normal, Aggressive\}$, where their specific meanings are illustrated in Table \ref{tab:Motion_Intention_Explain}.


\begin{table}
	\caption{Motion Intention Description}
	\label{tab:Motion_Intention_Explain}
	\begin{threeparttable}
		\setlength{\extrarowheight}{2.5pt}
		\begin{tabularx}{3.4in}{cX}
			\toprule
			Motion Intention& Description \\
			\midrule
			\midrule
			\textit{Stopping} & Obstacle vehicle will commit a full stop for giving way or parking\\
			\midrule
			\textit{Hesitating} & Obstacle vehicle will either decelerate or accelerate to adjust its driving behavior \\
			\midrule
			\textit{Normal} & Obstacle vehicle's behavior will comply with the generalized road context\\
			\midrule
			\textit{Aggressive} & Obstacle vehicle's behavior will be aggressive without any sense to negotiate \\
			\bottomrule
		\end{tabularx}
	\end{threeparttable}
\end{table}

Provided the generalized road context $\mathcal{C}$ and the inferred motion intention $\iota_{i}$, the state transition of obstacle vehicle $i$ is modeled as,
\begin{eqnarray}
\nonumber
\mathrm{Pr}(s^{[i]}_{t+\Delta t}|s^{[i]}_{t})=\mathrm{Pr}(x^{[i]}_{t+\Delta t},\iota^{[i]}_{t + \Delta t}|x^{[i]}_{t},\iota^{[i]}_{t})
\\
=\sum_{c(x^{[i]}_{t})} \mathrm{Pr}(x^{[i]}_{t+\Delta t},\iota^{[i]}_{t + \Delta t}|x^{[i]}_{t},\iota^{[i]}_{t},c(x^{[i]}_{t}))\mathrm{Pr}(c(x^{[i]}_{t})|x^{[i]}_{t}),
\label{eqn:Obst_State_Transition}
\end{eqnarray}
where $x^{[i]} =[\rm x^{[i]}, \rm y^{[i]}, \theta^{[i]}, \rm v^{[i]}]^{\rm T}$ is defined as the vehicle metric state for notation clarity, and $c(x^{[i]}_{t})\in\mathcal{C}$ denotes the road context information corresponding to vehicle state $x^{[i]}_{t}$ at time $t$. In detail, the road context $c(x^{[i]}_{t})$ features the possible driving directions according to the traffic rules and the typical (or reference) vehicle speed as well. As such, the obstacle vehicle state transition is driven by the inferred motion intention and constrained by the road context. 

While Eqn. (\ref{eqn:Obst_State_Transition}) can provide an reasonable modeling of the obstacle vehicle's state transition, the dependencies within the vehicle behaviors are not considered. Therefore, we seek to reformulate the state transition and make it conditional on a joint metric state $X^{[i]} = x^{[e]}\bigcup_{\forall j\in (\varphi_{\text{vis}}\setminus i)}x^{[j]}$ as,
\begin{eqnarray}
\nonumber
\mathrm{Pr}(x^{[i]}_{t+\Delta t},\iota^{[i]}_{t + \Delta t}|x^{[i]}_{t},\iota^{[i]}_{t},c(x^{[i]}_{t}))\\
=\sum_{X^{[i]}_{t}}\mathrm{Pr}(x^{[i]}_{t+\Delta t},\iota^{[i]}_{t + \Delta t}|x^{[i]}_{t},\iota^{[i]}_{t},c(x^{[i]}_{t}),X^{[i]}_{t})\mathrm{Pr}(X^{[i]}_{t}),
\label{eqn:Obst_State_Transition_Extended}
\end{eqnarray}
where the joint metric state $X^{[i]}_{t}$ is assumed to be conditional independent on the vehicle $i$'s state and the corresponding road context. The inclusion of $X^{[i]}_{t}$ into the state transition enables us to model the implicit vehicle dependencies as \emph{ each vehicle has to avoid the potential collision with any other vehicles}. Due to the sensing occlusion, the vehicle dependencies can only considered for the obstacle vehicles that are not blind to the ego vehicle. In a sense, the size of the joint metric state $X^{[i]}_{t}$ is a function of the sensing range. Therefore, the larger sensing range can be extended for ego vehicle, the more comprehensive vehicle behavior dependency can be considered. This is where cooperative perception can come into effect and contribute.

Referring to Eqn. (\ref{eqn:Obst_State_Transition_Extended}), we can always marginalize the transition function over an implicit obstacle vehicle action $a_{\rm obst} = [v_{\rm obst},\phi_{\rm obst}]^{\rm T}$, including speed $v_{\rm obst}$ and steering angle $\phi_{\rm obst}$, to explicitly model the obstacle vehicle state transition as,
\begin{eqnarray}
\nonumber
\mathrm{Pr}(x^{[i]}_{t+\Delta t},\iota^{[i]}_{t + \Delta t}|x^{[i]}_{t},\iota^{[i]}_{t},c(x^{[i]}_{t}),X^{[i]}_{t})
\\
\nonumber
= \sum_{{a_{\rm obst}}^{[i]}_{t}} \mathrm{Pr}(x^{[i]}_{t+\Delta t}|x^{[i]}_{t},{a_{\rm obst}}^{[i]}_{t})\mathrm{Pr}(\iota^{[i]}_{t + \Delta t}|\iota^{[i]}_{t})
\\
\times \mathrm{Pr}({a_{\rm obst}}^{[i]}_{t}|x^{[i]}_{t},\iota^{[i]}_{t},c(x^{[i]}_{t}),X^{[i]}_{t}) ,
\label{eqn:Obst_State_Transition_Detial}
\end{eqnarray}
where $\mathrm{Pr}(x^{[i]}_{t+\Delta t}|x^{[i]}_{t},{a_{\rm obst}}^{[i]}_{t})$ can follow the same transition model as that of ego vehicle given the inferred obstacle vehicle action $a_{\rm obst}$, and the motion intention remains constant within the transition process $\mathrm{Pr}(\iota^{[i]}_{t + \Delta t}|\iota^{[i]}_{t})$. 

Regarding the inference of the obstacle vehicle action $\mathrm{Pr}({a_{\rm obst}}^{[i]}_{t}|x^{[i]}_{t},\iota^{[i]}_{t},c(x^{[i]}_{t}),X^{[i]}_{t})$, the obstacle vehicle speed can be properly modeled using the inferred motion intention and the implicit vehicle dependencies. The steering angle, however, is not featured by the motion intention, thus we aim at using the road context to enable the steering angle prediction. As such, the obstacle vehicle action inference can be reformulated as,
\begin{eqnarray}
\nonumber
\mathrm{Pr}({a_{\rm obst}}^{[i]}_{t}|x^{[i]}_{t},\iota^{[i]}_{t},c(x^{[i]}_{t}),X^{[i]}_{t})
\\
=\mathrm{Pr}({v_{\rm obst}}^{[i]}_{t}|\iota^{[i]}_{t},X^{[i]}_{t})\mathrm{Pr}({\phi_{\rm obst}}^{[i]}_{t}|c(x^{[i]}_{t})).
\label{eqn:Obst_Speed_Steering_Inference}
\end{eqnarray}
In short, the speed inference $\mathrm{Pr}({v_{\rm obst}}^{[i]}_{t}|\iota^{[i]}_{t},X({a_{\text{\tiny CPoD}}}_{t}))$ is accomplished by mapping the acceleration command to the inferred motion intention $\iota$ according to Table \ref{tab:Motion_Intention_Explain}, meanwhile the obstacle vehicle can still have certain chance to commit a deceleration if the collision risk with any other vehicles is high. The inference over the steering control is more straightforward, which is sampled from the possible driving directions according to the traffic rules featured by the road context $c(x^{[i]}_{t})$. The reader can refer to \cite{liu2015situation} for a more detailed discussion.

\subsubsection{Observation Modeling}
Thanks to the recent research advances in vehicle detection and tracking, the vehicle metric state $x$ can be properly observed with Gaussian noise imposed, thereby we can simply generate the observations with an one-on-one mapping directly from the corresponding metric states. As such, the observation of any vehicle $i$ is modeled as an vector consisting of the values of vehicle pose and velocity. The joint observation $z$ thereby is defined as $z = z^{[e]}\bigcup_{\forall i\in\varphi_{\text{vis}}}z^{[i]}$. In a sense, only the vehicles that are falling inside the extended sensing map $\mathbf{Z}^{[e]}$ are observed.

\subsubsection{Belief Tracking}
Since the obstacle vehicle motion intention can only be partially observable, the motion intention needs to be maintained as a belief state. Given the augmented observation, the belief tracker is dedicated for inferring and tracking the obstacle vehicle motion intentions using Bayes rules. The inference of obstacle vehicle $i$'s motion intention thereby can be formulated as,
\begin{equation}
 b(\iota^{[i]}_{t+\Delta t}) = \eta\mathbb{
 	P}(\iota^{[i]}_{t+\Delta t}|\iota^{[i]}_{t})\mathrm{Pr}(z^{[i]}_{t+\Delta t}|\iota^{[i]}_{t+\Delta t})b(\iota^{[i]}_{t}),
 \end{equation}
where $\eta$ is the normalization factor, and the motion intention remains constant within the transition process. 


Regarding the observation function $\mathrm{Pr}(z^{[i]}|\iota^{[i]})$, we model it as a Gaussian with mean $\mathbf{m}(z^{[i]}\ominus c(z^{[i]}), \iota^{[i]})$ and covariance $\mathbf{\Sigma}(c(z^{[i]}))$, where the operator $\ominus$ calculates the deviation of the observed vehicle state $z^{[i]}$ from the corresponding road context $c(z^{[i]})$. Recall our earlier discussion, we aim at using this deviation as an interesting hint to infer the motion intention. More specifically, the speed deviation is employed in this study, where the observed vehicle speed is given as $\rm v^{[i]}$, and the road context $c(z^{[i]})$ provides the reference speed $\rm v_{\rm ref}(z^{[i]})$ that is generalized from the vehicle behavior training data using Gaussian Process\cite{liu2015roadrule}. Thereafter, the mean value $\mathbf{m}(z^{[i]}\ominus c(z^{[i]}), \iota^{[i]})$ can be reformulated as $\mathbf{m}(\rm v^{[i]}\ominus \rm v_{\rm ref}(x^{[i]}), \iota^{[i]})$ and defined in Table \ref{tab:Motion_Intention_Model}, which is proposed according to the motion intention definition in Table \ref{tab:Motion_Intention_Explain}. Moreover, the covariance function $\mathbf{\Sigma}(c(z^{[i]}))$ is defined as $\mathbf{\Sigma}(c(z^{[i]})) = \rm v_{\rm ref}(z^{[i]})/\sigma$, where the scaling factor $\sigma$ is to adjust the belief confidence. 

\begin{table}
	\caption{Mean function w.r.t. Motion Intention}
	\centering
	\setlength{\tabcolsep}{.4em}
	\def\arraystretch{1.3}
	\label{tab:Motion_Intention_Model}    
	\setlength{\extrarowheight}{2.5pt}
	\begin{threeparttable}
		\begin{tabular}{c|cccc}
			\hline
			$\iota\in\mathcal{I}$ & $Stopping$ & $Hesitating$ & $Normal $ &  $Aggressive$ \\
			\hline
			$\mathbf{m}(\rm v\ominus \rm v_{\rm ref}(z),\iota)$&  $0.0$ & $\mathrm{v}_{\mathrm{ref}}(z)/2$ & $\mathrm{v}_{\mathrm{ref}}(z)$ & $3\mathrm{v}_{\mathrm{ef}}(z)/2$ \\
			\hline
		\end{tabular}
	\end{threeparttable}
\end{table}
 

\subsubsection{Reward Function}
The main objective of the ego vehicle is to arrive the destination as quickly as possible, meanwhile avoid collision with any other obstacle vehicles. Within this process, the communication within the vehicles needs to be actively requested if the situation becomes tricky. Two sub-decisions have to be made by an ego vehicle, thus the reward function $R(s,a):\mathcal{S}\times\mathcal{A}\rightarrow\mathbb{R}$ is decoupled as,
\begin{equation}
R(s,a) = R(s,a_{\text{\tiny ACC}}) + R(s,a_{\text{\tiny CPoD}}).
\end{equation}

For acceleration-based reward function $R(s,a_{\text{\tiny ACC}}):\mathcal{S}\times\mathcal{A}_{\text{\tiny ACC}}\rightarrow\mathbb{R}$, we want the vehicle move safely, efficiently and smoothly, thus it is defined as $R(s,a_{\text{\tiny ACC}}) = R_{\text{goal}}(s,a_{\text{\tiny ACC}}) + R_{\text{crash}}(s,a_{\text{\tiny ACC}})+ R_{\text{action}}(a_{\text{\tiny ACC}}) + R_{\text{speed}}(s)$. The reward function $R_{\text{goal}}(s,a_{\text{\tiny ACC}})$ provides a reward when the vehicle reach the destination, and $R_{\text{crash}}(s,a_{\text{\tiny ACC}})$ outputs a high penalty if the ego vehicle is in collision. The action reward $R_{\text{action}}(a_{\text{\tiny ACC}})$ targets to smooth the vehicle velocity by proving a small penalty if $a_{\text{\tiny ACC}}\neq\textsc{Constant}$. The speed reward $R_{\text{speed}}(s) = k\rm v/\rm v_{max}$ is to encourage high speed travel and improve the driving efficiency, where $\rm v_{max}$ is the ego vehicle's maximum speed and $k$ is a scaling factor.

Regarding CPoD based reward function $R(s, a_{\text{\tiny CPoD}}):\mathcal{S}\times\mathcal{A}_{\rm CPoD}\rightarrow\mathbb{R}$, we want to emphasize the cost of communication by introducing a constant penalty $R_{\text{comm.}}(a_{\text{\tiny CPoD}})$ every time cooperative perception is activated. Once the situation awareness becomes tricky or the collision risk is high, the cooperative perception should be activated to gain more sensing information for proper decision making. As such, we introduce the $R_{\text{intention}}(s)$ to reward cooperative perception if the IOV's motion intention $\iota^{[iov]}\neq\textsc{Normal}$, in which case IOV's behavior becomes harder to predict. Moreover, another reward $R_{\text{\tiny TTC}}(s,a_{\text{\tiny CPoD}})$ will be imposed once the Time-To-Collision between ego vehicle and IOV is lower than the designed threshold. In summary, the $\rm CPoD$-driven reward function is formulated as $R(s, a_{\text{\tiny CPoD}}) = R_{\text{comm.}}(a_{\text{\tiny CPoD}}) + R_{\text{intention}}(s) + R_{\text{\tiny TTC}}(s,a_{\text{\tiny CPoD}})$.

Worth mentioning that we design the reward function $R(s, a_{\text{\tiny CPoD}})$ w.r.t. the vehicle state rather than the belief state, the purpose is to ensure that the POMDP value function stays piece-wise linear and convex, although the usage of belief state or entropy to evaluate information gain is more reasonable in some cases. 

\subsection{POMDP Solver}
Given the designed POMDP model, the online POMDP solver DESPOT is employed for its efficiency of handling the large observation space \cite{somani2013despot}. As an online POMDP solver, DESPOT only searches the belief space that is achievable from the current belief state and interleaves the planning and executing phases, Moreover, rather than searching the whole belief tree, DESPOT only samples a \textit{scenario} set with constant size $Q$. As a consequence, the belief tree of height $H$ contains only $O(|\mathcal{A}|^{H}Q)$ nodes. The belief state is represented by the random particles within DESPOT, and for each particle the obstacle vehicles' motion intentions are randomly sampled for belief state representation.

\section{Evaluation}\label{sec:Evaluation}
In this section, the proposed algorithm will be evaluated to validate its functionality.

\subsection{Settings}

\begin{figure}
	\begin{center}
		\includegraphics[width=3.4in]{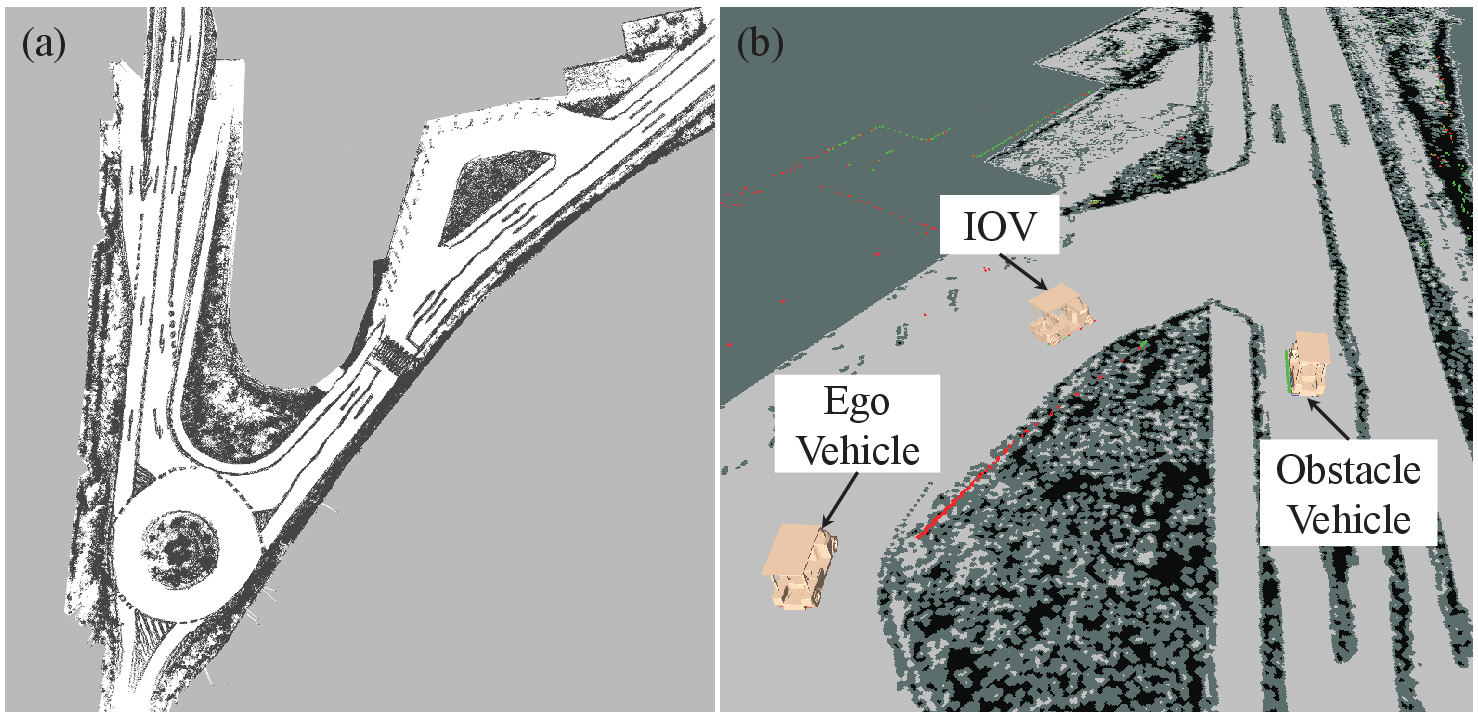}
		\caption{Evaluation environment and scenario: (a) shows the evaluation environment, which is a typical urban road within the campus of National University of Singapore. (b) depicts the evaluation scenario, where the red points represent the ego vehicle's laser scan and the IOV's laser scan is shown by the green points. }
		\label{fig:Environment_Scenario}
	\end{center}
\end{figure}

Given the environment and the corresponding road context as Fig. \ref{fig:Environment_Scenario}(a), the \textit{Stage} simulator is employed for simulation evaluation, where Gaussian noise is purposely imposed on the vehicle pose and velocity measurement. 

Implementation-wise, each vehicle is running an independent navigation system in ROS \cite{ROS2009Citation}, where the vehicle communication is accomplished by sharing an unique ROS core. The proposed algorithm is running with 2 Hz for both CPoD and vehicle acceleration control, and the Pure-Pursuit tracking algorithm is employed for the vehicle steering control \cite{chong2013autonomy}.



\begin{figure*}
	\begin{center}
		\includegraphics[width=6.8in]{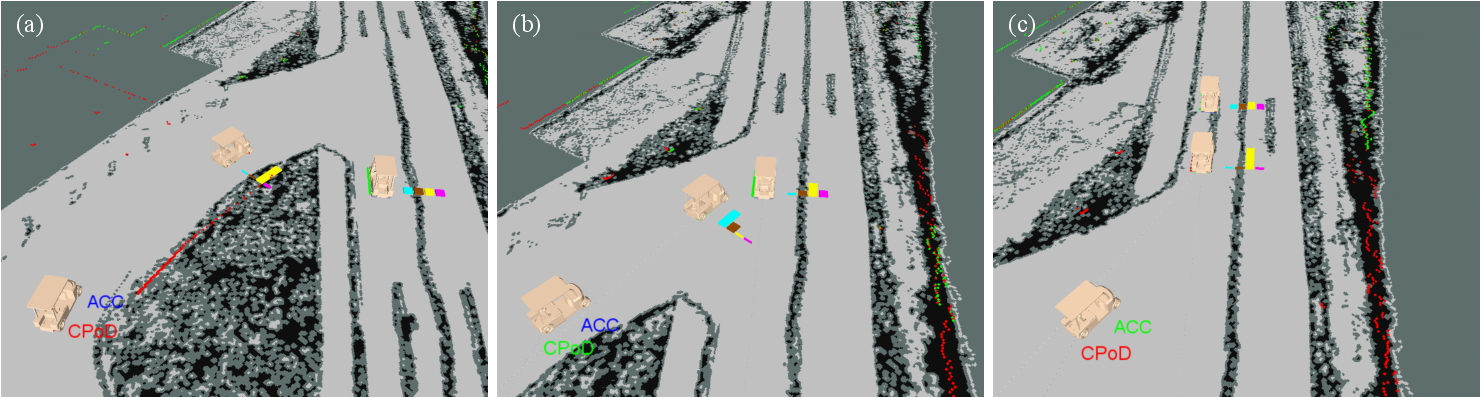}
		\caption{Navigation process of situation-aware decision making with CPoD at T-junction. The actions are highlighted with the colorful text. For acceleration decision:  [Red: $\textsc{Decelerate}$, Green: $\textsc{Accelerate}$, Blue: $\textsc{Constant}$]. For CPoD decision: [Red: $\textsc{Deactive}$, Green: $\textsc{Active}$] The motion intention is represented as the cubic marker: [Sky Blue: $Stopping$, Brown: $Hesitating$, Yellow: $Normal$, Purple: $Aggressive$], where the cubic height is proportional to the corresponding belief value. }
		\label{fig:CP_Tjunction_Negotiate}
	\end{center}
\end{figure*}

\subsection{Results}
The proposed algorithm has been successfully applied for several urban road driving scenarios within the evaluation stage. Here, we want to highlight the ability of our algorithm to address the autonomous driving decision making at the T-junction (see Fig. \ref{fig:Environment_Scenario}(b)), where the evaluating scenario is same as what is discussed in Section \ref{sec:Problem_Statement}.

As illustrated in Fig. \ref{fig:CP_Tjunction_Negotiate}(a), the ego vehicle is following the leading vehicle (identified as IOV) to approach the T-junction for lane merging. Meanwhile, there comes another obstacle vehicle that is blind to the ego vehicle, which means its motion intention is quite ambiguous because no observation is available. Before the IOV and the obstacle vehicle getting interacted, the ego vehicle reasonably maintains a smooth speed to follow the IOV while deactivating the CPoD to save the "expensive" communication. This balance breaks down when the IOV commits a deceleration in order to give way to the obstacle vehicle. Right after the IOV's motion intention is updated, the ego vehicle actively enables the CPoD to extend the sensing range and start to infer the obstacle vehicle's motion intention as Fig. \ref{fig:CP_Tjunction_Negotiate}(b). Given the inferred situation evolution, especially the behavior correlation between IOV and obstacle vehicle, the ego vehicle decides to decelerate far before the obstacle clearance become critical, this is thanks to the cooperative perception which makes the vehicle behavior dependency analysis become achievable. After the obstacle vehicle becoming clear, the IOV starts to move forward for lane merging, the ego vehicle also decides to speed up while keeping the CPoD being deactivated (see Fig. \ref{fig:CP_Tjunction_Negotiate}(c)). 

\begin{figure}
	\begin{center}
		\includegraphics[width=3.4in]{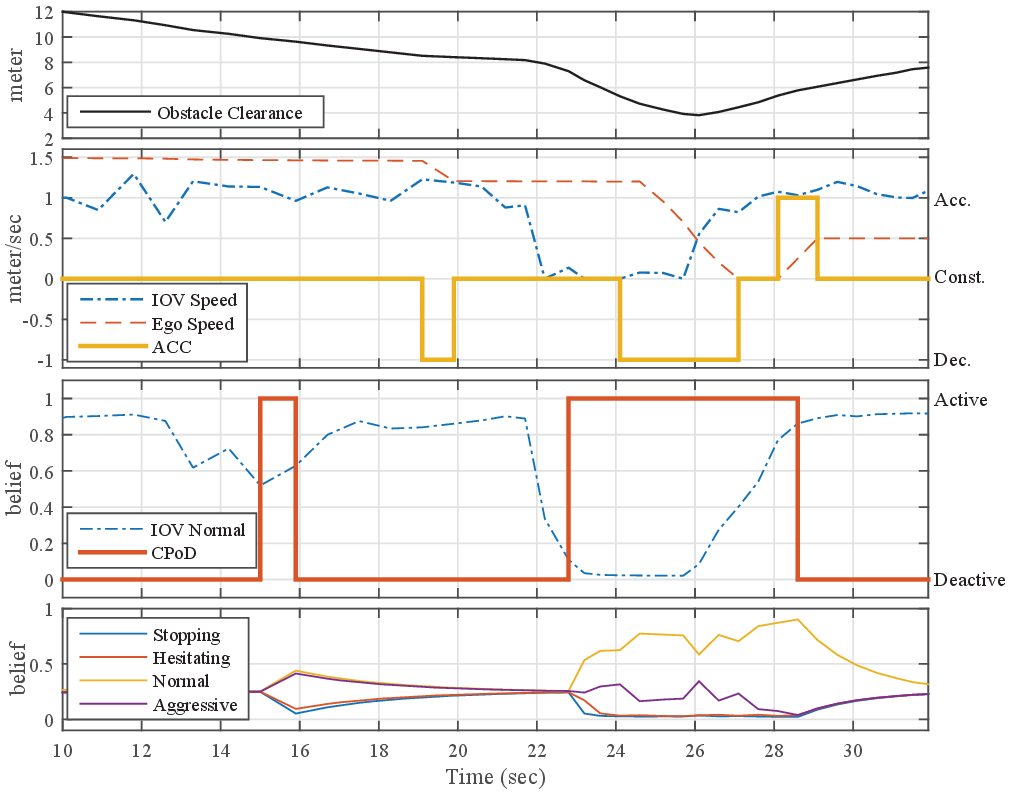}
		\caption{Situation evolution analysis: (a) shows the obstacle clearance and (b) represents the acceleration command along with the speeds of both IOV and ego vehicle. (c) depicts the CPoD decision with the IOV's \textsc{Normal} belief, and (d) illustrates the obstacle vehicle's motion intention belief. }
		\label{fig:Decision_Analysis}
	\end{center}
\end{figure}

The detailed illustration of the situation evolution can be found in Fig. \ref{fig:Decision_Analysis}, where Fig. \ref{fig:Decision_Analysis}(a) depicts the obstacle clearance. The acceleration decision, along with the speed of both ego vehicle and IOV, are shown in Fig. \ref{fig:Decision_Analysis}(b), from which we can find that acceleration decision is quite responsive to the obstacle clearance and the driving efficiency. Moreover, the CPoD decision and the IOV's motion intention is shown in Fig. \ref{fig:Decision_Analysis}(c), where only the \textsc{Normal} belief is plotted because it is directly related to the CPoD decision. An interesting observation is that the CPoD decision is highly depends on the IOV's motion intention, which is exactly our expected behavior. In order to demonstrate the CPoD's impact on the obstacle vehicle, we also plot the obstacle vehicle's motion intention in Fig. \ref{fig:Decision_Analysis}(d), where the motion intention belief can be quickly updated when the CPoD is activated, otherwise the motion intention will stay ambiguous. 

Acknowledging that the proposed approach is probabilistic in nature, we extensively conducted 100 evaluation trials for this T-junction scenario. For the sake of investigating the impact of cooperative perception on decision making, we also implemented an algorithm with the CPoD being deactivated constantly and we name it as CPWO. Moreover, in order to address the concern that the introduction of the CPoD might make the decision sub-optimal compared to the case where cooperative perception is always activated, another algorithm, CPW, with CPoD being activated all the time is implemented as well. 

\begin{figure}
	\begin{center}
		\includegraphics[width=3.4in]{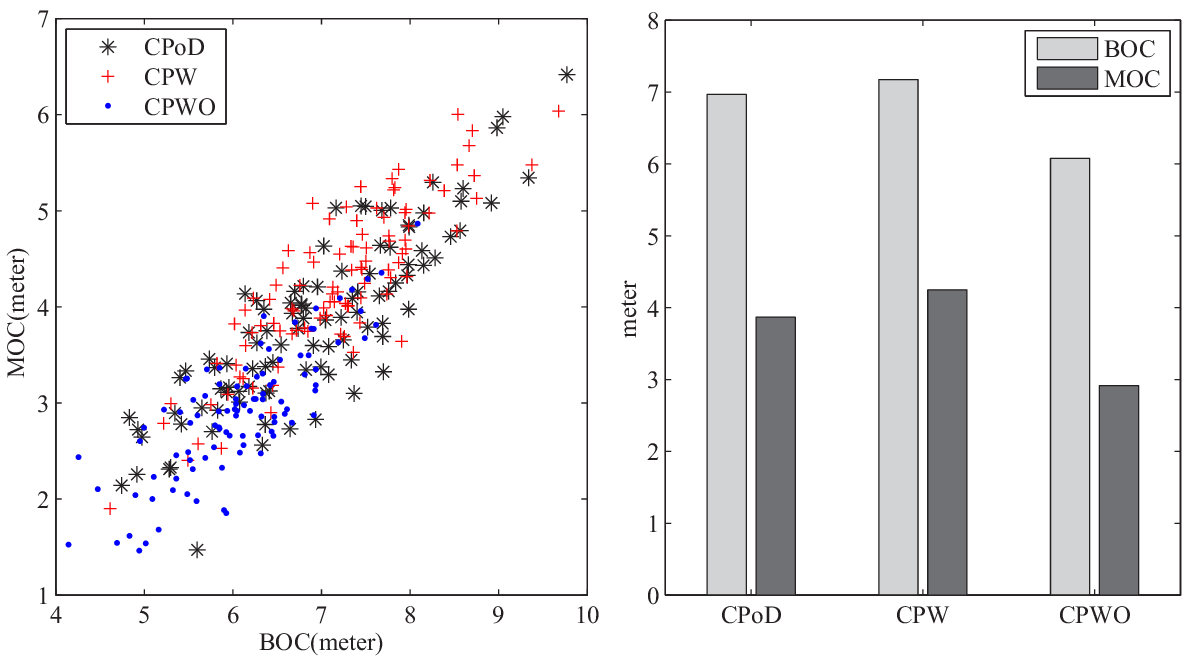}
		\caption{Navigation performance comparison w.r.t. BOC and MOC: The evaluation results are depicted in (a) and the corresponding averaged values are shown in (b).}
		\label{fig:Comparizon_Result}
	\end{center}
\end{figure}

In order to compare the decision making performance, two evaluating metrics are proposed, which includes the Braking Obstacle Clearance (BOC) and Minimal Obstacle Clearance (MOC). More specifically, when the IOV is stopping for the coming obstacle vehicle, the BOC represents the distance between the ego vehicle and IOV when the ego vehicle decides to decelerate. Moreover, MOC measures the minimal distance between the ego vehicle and the IOV during the whole navigation process. The results are depicted in Fig. \ref{fig:Comparizon_Result}, where the BOC and MOC are plotted in two separate dimensions in Fig. \ref{fig:Comparizon_Result}(a) and the corresponding mean values are represented in Fig. \ref{fig:Comparizon_Result}(b). By comparing the navigation performance of CPoD and CPWO, we can safely conclude that the inclusion of cooperative perception can helpfully improve the driving safety, because the ego vehicle is able to react earlier and maintain a larger safety gap with the leading vehicle. Moreover, the CPoD can achieve almost the same performance as that of CPW, which hints that the modeling of CPoD inside POMDP can properly identify the situation where cooperative perception is necessary, whereas the communication constraints can be alleviated. 

\begin{figure*}
	\begin{center}
		\includegraphics[width=6.8in]{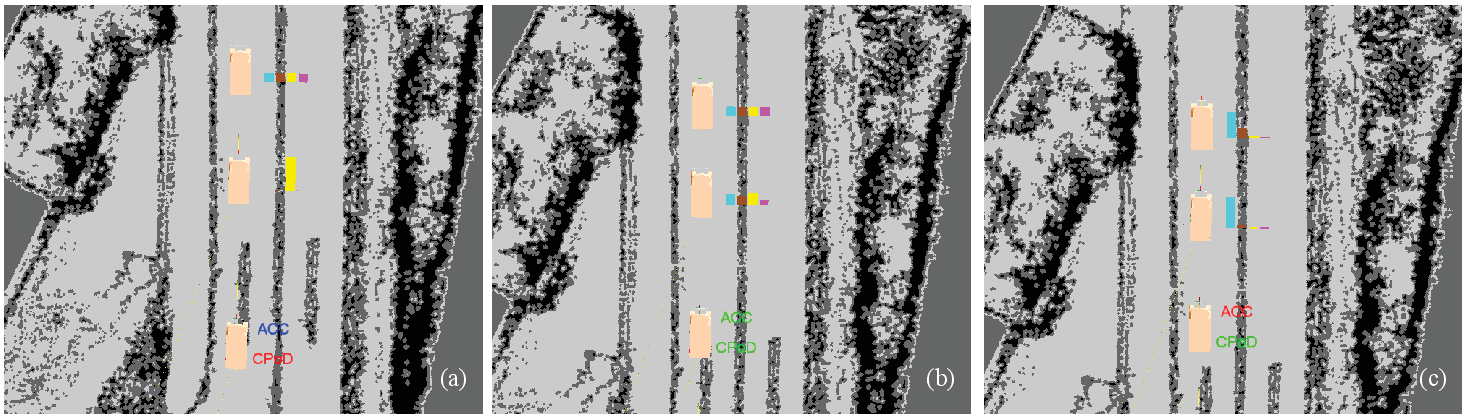}
		\caption{Navigation process of situation-aware decision making with CPoD on a single-lane road (Top-down view. }
		\label{fig:CP_SingleLane_Negotiate}
	\end{center}
\end{figure*}

Given the promising result we achieved for lane-merging at T-junction, we extend the evaluation scenario to an single-lane road where the ego vehicle is following a leading vehicle that blocks its sight-view in Fig. \ref{fig:CP_SingleLane_Negotiate}(a). Similarly, the obstacle vehicle is not visible to the ego vehicle, such that the ego vehicle need to reasonably active the CPoD and choose the optimal acceleration command to drive safely and efficiently. The overall navigation result is demonstrated in Fig. \ref{fig:CP_SingleLane_Negotiate}, which shows similar performance as that of the T-junction scenario. (a) shows the evaluating scenario, where the most front obstacle vehicle is blind to the ego vehicle. The most front obstacle vehicle suddenly commits a stop and the leading vehicle starts to decelerate as well, the CPoD is then activated accordingly as in (b). Thereafter, the ego vehicle safely commit an deceleration while making the CPoD be enabled in (c).


\section{Conclusion}\label{sec:Conclusion}
In this paper, we introduced a problem of situation-aware autonomous driving decision making with cooperative perception. The extended perception range contributed by the cooperative perception is properly employed to address the implicit dependencies within the vehicles. Meanwhile, we acknowledge the limitation of wireless communication and propose a CPoD scheme. The situation-aware decision making problem, together with the CPoD, is unified as a POMDP model and solved in an online manner. The extensive evaluations show that the proposed algorithm can properly improve the driving functionality and the perception sharing. The future work can be contributed to improving the identification of IOV and extending the simulation evaluations to the real experiments as well.

\bibliographystyle{IEEEtran}
\bibliography{ICRA}

\begin{thebibliography}{10}
\providecommand{\url}[1]{#1}
\csname url@rmstyle\endcsname
\providecommand{\newblock}{\relax}
\providecommand{\bibinfo}[2]{#2}
\providecommand\BIBentrySTDinterwordspacing{\spaceskip=0pt\relax}
\providecommand\BIBentryALTinterwordstretchfactor{4}
\providecommand\BIBentryALTinterwordspacing{\spaceskip=\fontdimen2\font plus
\BIBentryALTinterwordstretchfactor\fontdimen3\font minus
  \fontdimen4\font\relax}
\providecommand\BIBforeignlanguage[2]{{%
\expandafter\ifx\csname l@#1\endcsname\relax
\typeout{** WARNING: IEEEtran.bst: No hyphenation pattern has been}%
\typeout{** loaded for the language `#1'. Using the pattern for}%
\typeout{** the default language instead.}%
\else
\language=\csname l@#1\endcsname
\fi
#2}}

\bibitem{sivaraman2013observing}
S.~Sivaraman, B.~Morris, and M.~Trivedi, ``Observing on-road vehicle behavior:
  Issues, approaches, and perspectives,'' in \emph{Intelligent Transportation
  Systems-(ITSC), 2013 16th International IEEE Conference on}.\hskip 1em plus
  0.5em minus 0.4em\relax IEEE, 2013, pp. 1772--1777.

\bibitem{bandyopadhyay2013intention}
T.~Bandyopadhyay, C.~Z. Jie, D.~Hsu, M.~H. Ang~Jr, D.~Rus, and E.~Frazzoli,
  ``Intention-aware pedestrian avoidance,'' in \emph{Experimental
  Robotics}.\hskip 1em plus 0.5em minus 0.4em\relax Springer International
  Publishing, 2013, pp. 963--977.

\bibitem{li2012cooperative}
H.~Li, ``Cooperative perception: Application in the context of outdoor
  intelligent vehicle systems,'' Ph.D. dissertation, Paris, ENMP, 2012.

\bibitem{kim2014multivehicle}
S.-W. Kim, B.~Qin, Z.~J. Chong, X.~Shen, W.~Liu, M.~Ang, E.~Frazzoli, and
  D.~Rus, ``Multivehicle cooperative driving using cooperative perception:
  Design and experimental validation,'' \emph{IEEE Transaction on Intelligent
  Transportation System}, 2014.

\bibitem{liu2015situation}
W.~Liu, S.-W. Kim, and M.~H. Ang, ``Situation-aware decision making for
  autonomous driving on urban road using online {POMDP},'' in \emph{Intelligent
  Vehicles (IV), 2015 IEEE Symposium on}.\hskip 1em plus 0.5em minus
  0.4em\relax IEEE, 2015.

\bibitem{liu2013motion}
W.~Liu, S.~Kim, Z.~Chong, X.~Shen, and M.~Ang, ``Motion planning using
  cooperative perception on urban road,'' in \emph{Robotics, Automation and
  Mechatronics (RAM), 2013 6th IEEE Conference on}.\hskip 1em plus 0.5em minus
  0.4em\relax IEEE, 2013, pp. 130--137.

\bibitem{liu2014vehicle}
W.~Liu, S.-W. Kim, K.~Marczuk, and M.~H. Ang, ``Vehicle motion intention
  reasoning using cooperative perception on urban road,'' in \emph{Intelligent
  Transportation Systems (ITSC), 2014 IEEE 17th International Conference
  on}.\hskip 1em plus 0.5em minus 0.4em\relax IEEE, 2014, pp. 424--430.

\bibitem{spaan2009decision}
M.~T. Spaan and P.~U. Lima, ``A decision-theoretic approach to dynamic sensor
  selection in camera networks.'' in \emph{ICAPS}, 2009.

\bibitem{spaan2008cooperative}
M.~T. Spaan, ``Cooperative active perception using {POMDPs},'' in \emph{AAAI
  2008 workshop on advancements in POMDP solvers}, 2008.

\bibitem{otte2014any}
M.~Otte and N.~Correll, ``Any-com multi-robot path-planning with dynamic teams:
  Multi-robot coordination under communication constraints,'' in
  \emph{Experimental Robotics}.\hskip 1em plus 0.5em minus 0.4em\relax
  Springer, 2014, pp. 743--757.

\bibitem{ferguson2008motion}
D.~Ferguson, T.~M. Howard, and M.~Likhachev, ``Motion planning in urban
  environments,'' \emph{Journal of Field Robotics}, vol.~25, no. 11-12, pp.
  939--960, 2008.

\bibitem{fletcher2008cornell}
L.~Fletcher, S.~Teller, E.~Olson, D.~Moore, Y.~Kuwata, J.~How, J.~Leonard,
  I.~Miller, M.~Campbell, D.~Huttenlocher, \emph{et~al.}, ``The mit--cornell
  collision and why it happened,'' \emph{Journal of Field Robotics}, vol.~25,
  no.~10, pp. 775--807, 2008.

\bibitem{ulbrich2013probabilistic}
S.~Ulbrich and M.~Maurer, ``Probabilistic online {POMDP} decision making for
  lane changes in fully automated driving,'' in \emph{Intelligent
  Transportation Systems}, 2013, pp. 2063--2070.

\bibitem{bai2015intention}
H.~Bai, S.~Cai, N.~Ye, D.~Hsu, and W.~S. Lee, ``Intention-aware online {POMDP}
  planning for autonomous driving in a crowd,'' in \emph{Robotics and
  Automation (ICRA), 2015 IEEE International Conference on}.\hskip 1em plus
  0.5em minus 0.4em\relax IEEE, 2015, pp. 454--460.

\bibitem{shaojun2014online}
C.~Shaojun, ``Online {POMDP} planning for vehicle navigation in densely
  populated area,'' 2014.

\bibitem{xiaotong2014dars}
X.~Shen, S.~Pendleton, and M.~H. Ang~Jr., ``Scalable cooperative localization
  with minimal sensor configuration,'' in \emph{International Symposium on
  Distributed Autonomous Robotics System}, 2014.

\bibitem{lavalle2006planning}
S.~M. LaValle, \emph{Planning algorithms}.\hskip 1em plus 0.5em minus
  0.4em\relax Cambridge university press, 2006.

\bibitem{liu2015roadrule}
W.~Liu, S.-W. Kim, and M.~H. Ang, ``Probabilistic road context inference for
  autonomous vehicles,'' in \emph{Robotics and Automation (ICRA), 2015 IEEE
  International Conference on}.\hskip 1em plus 0.5em minus 0.4em\relax IEEE,
  2015.

\bibitem{somani2013despot}
A.~Somani, N.~Ye, D.~Hsu, and W.~S. Lee, ``{DESPOT}: Online {POMDP} planning
  with regularization,'' in \emph{Advances In Neural Information Processing
  Systems}, 2013, pp. 1772--1780.

\bibitem{ROS2009Citation}
M.~Quigley, K.~Conley, B.~P. Gerkey, J.~Faust, T.~Foote, J.~Leibs, R.~Wheeler,
  and A.~Y. Ng, ``Ros: an open-source robot operating system,'' in \emph{ICRA
  Workshop on Open Source Software}, 2009.

\bibitem{chong2013autonomy}
Z.~Chong, B.~Qin, T.~Bandyopadhyay, T.~Wongpiromsarn, B.~Rebsamen, P.~Dai,
  E.~Rankin, and M.~H. Ang~Jr, ``Autonomy for mobility on demand,''
  \emph{Intelligent Autonomous Systems 12}, pp. 671--682, 2013.

\end{thebibliography}

\end{document}